\documentclass{INTERSPEECH2023}

\interspeechcameraready 
\usepackage{enumitem}
\usepackage[labelfont=bf]{caption}

\title{DisfluencyFixer: A tool to enhance Language Learning through Speech To Speech Disfluency Correction}
\name{Vineet Bhat, Preethi Jyothi, Pushpak Bhattacharyya}
\address{
  IIT Bombay, India}
\email{180260042@iitb.ac.in, pjyothi@cse.iitb.ac.in, pb@cse.iitb.ac.in}

\begin{document}

\maketitle
 
\begin{abstract}
Conversational speech often consists of deviations from the speech plan, producing disfluent utterances that affect downstream NLP tasks. Removing these disfluencies is necessary to create fluent and coherent speech. This paper presents DisfluencyFixer, a tool that performs speech-to-speech disfluency correction in English and Hindi using a pipeline of Automatic Speech Recognition (ASR), Disfluency Correction (DC) and Text-To-Speech (TTS) models. Our proposed system removes disfluencies from input speech and returns fluent speech as output along with its transcript, disfluency type and total disfluency count in source utterance, providing a one-stop destination for language learners to improve the fluency of their speech. We evaluate the performance of our tool subjectively and receive scores of 4.26, 4.29 and 4.42 out of 5 in ASR performance, DC performance and ease-of-use of the system. Our tool can be accessed openly at the following link\footnote{\protect\url{https://www.cfilt.iitb.ac.in/speech2text/}}.

\end{abstract}

\section{Introduction}

Speech-based interactive features are rapidly expanding into appliances such as AI assistants, chatbots, autonomous driving, customer service, etc. However, they are challenging due to variations in speech, such as dialects, pronunciations and impairments. In addition, humans often do not have a well-defined speech plan that introduces filler words, repetition or correction of uttered phrases, making it difficult for the downstream understanding of the spoken utterance \cite{gupta-etal-2021-disfl}. Words that are part of the spoken utterance but do not add meaning to the sentence are termed disfluencies, and removing them becomes an essential post-processing task for speech technologies \cite{Honal2003CorrectionOD}. 

Previous work has used ASR in English language learning for foreign speakers \cite{morton_interactive_2012}. TTS has been applied for language learning by providing a method to aid learners in understanding correct pronunciations and prosody of the English language \cite{Cardoso2015EvaluatingTS}. In this work we propose DisfluencyFixer, a tool that integrates ASR, DC and TTS models for performing Speech To Speech Disfluency Correction. In contrast to existing related technologies for speech-based language learning, we extend our tool to two languages - English and Hindi and provide an easy-to-use interface for deploying the system in any noisy environment.

\section{DisfluencyFixer: A tool for Speech To Speech Disfluency Correction}

DisfluencyFixer is an online web service created using ReactJS frontend and Python backend. It consists of an interactive user interface for automatic disfluency correction in input speech generating fluent output speech with transcription, information about the total disfluency count and type of disfluency. We describe the salient features of our tool below. 

\begin{figure*}[t]
\centering
    \includegraphics[scale=0.30]{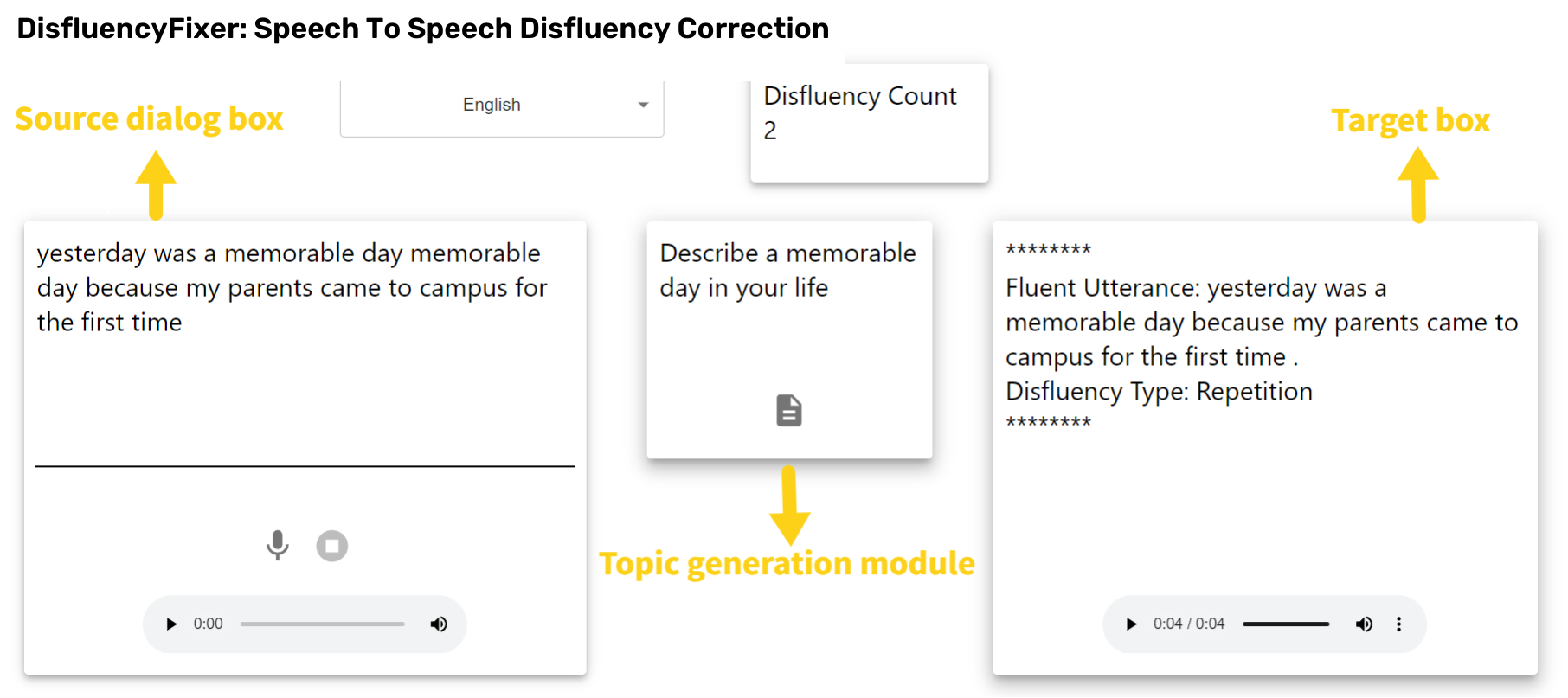}
    \caption{Screenshot of DisfluencyFixer tool}\label{fig:a}
\end{figure*}

\subsection{Features}
\begin{enumerate}[noitemsep, nosep, leftmargin=*,label=\textbf{\arabic*.}]
\item \textbf{User Interface:} We provide a simple drop-down button to select between the two languages for input. The interface has two main components, as shown in Figure \ref{fig:a}. The first component is the source dialog box, allowing users to start or stop recording their speech for analysis and fluent speech generation. They can also view the raw ASR transcript of their spoken utterance along with the audio file to hear their input. The second component is the target box, which shows the fluent transcript of the utterance and the type of disfluency identified. Users can also directly hear the fluent speech as output in this box. 
\item \textbf{Topic generation module:} This feature allows the user to generate a random prompt for the conversation, mimicking the environment of standard English learning exams like TOEFL or IELTS. After reading the prompt, users can speak into the tool using the source dialog box.
\item \textbf{ASR Module:} Our backend ASR is the XLSR-wav2vec 2.0 \cite{conneau_xlsr} model finetuned on the CommonVoice \cite{ardila-etal-2020-common} English and Hindi datasets. We chose this model because of its low latency and word error rate in noisy settings. 
\item \textbf{DC Module:} Our DC system is based on the MuRIL  transformer \cite{Khanuja2021MuRILMR} architecture and trained on annotated disfluent English sentences from the Switchboard corpus \cite{Godfrey1992SWITCHBOARDTS}. For Hindi, we train the system in a zero-shot setting using synthetically generated data \cite{kundu-etal-2022-zero}. Since MuRIL is pretrained on many Indian languages, the representations it produces are of high quality. Finetuning the model for token classification enables it to classify every word in the sentence as disfluent or fluent. The final fluent text is obtained by removing all the disfluent words marked by the model. 
\item \textbf{TTS Module:} We train a Forward Tacotron model\footnote{\protect\url{https://github.com/as-ideas/ForwardTacotron}} with a CARGAN \cite{Morrison2021ChunkedAG} vocoder on the IndicTTS \footnote{\protect\url{https://www.iitm.ac.in/donlab/tts/}} dataset in English and Hindi and deploy it for TTS synthesis. 
\item \textbf{Disfluency Classifier:} After performing manual annotation of real and synthetic data from \cite{kundu-etal-2022-zero}, we finetune MuRIL transformer \cite{Khanuja2021MuRILMR} for multi-class classification using a softmax classifier over the following disfluency types: Filler, Repetition, Correction, False Start and Fluent. 
\item \textbf{Disfluency Count Indicator:} The disfluency count indicator on the top right side of the interface measures the number of words removed from source utterance to create fluent speech. A higher disfluency count implies incoherent speech providing a way of automatic user feedback.

\end{enumerate}

\subsection{User Study}

We asked 48 language learners to evaluate our tool on three parameters - Performance of ASR, Performance of DC and Ease of Use. Of these 48 participants, 30 were students potentially preparing for language testing exams. The remaining surveyors were language learners, either in English or Hindi. The results of the survey are described in Table \ref{tab:eval} below. 

\begin{table}[h]
        \centering
        \setlength{\belowcaptionskip}{5pt} 
        \begin{tabular}{ | p{3cm}| c|}
        \hline
            \centering Evaluation Metric & Score (out of 5.0) \\
            \hline
            \centering Performance of ASR & 4.26\\
            \hline
            \centering Performance of DC & 4.29 \\
            \hline
            \centering Ease of Use
             & 4.42 \\
        \hline
        \end{tabular}
        \caption{Results of our subjective survey to evaluate the DisfluencyFixer tool; We report average score across both languages for the first two evaluation metrics.} \label{tab:eval}
        
    \end{table}

\vspace{-20pt}

The survey demonstrates scope of improvement for our ASR and DC models. Our ASR model makes phonetic mistakes that are propagated to the DC model. An interesting observation is that our tool performs better when users test it on their laptop/desktop than on mobile phones/tablets, implying that the tool performance correlates significantly with the type of microphone used to record speech. 

\section{Conclusion and Future Work}

In this paper, we introduce DisfluencyFixer, a tool which uses disfluency identification and correction to aid language learners improve the fluency of their speech. Our designed interface integrates ASR and TTS models to allow users to interact with the system through speech. The raw transcripts of disfluent and fluent speech, disfluency count and its type provide users with feedback on their speech and how to improve their conversational skills. 

We are working on improving our backend ASR and DC models to improve the tool's overall performance. We also plan to add more insights and metrics in the web service to improve user experience based on the feedback we received from our survey.

\bibliographystyle{IEEEtran}
\bibliography{mybib}

\end{document}